\documentclass{article}
\usepackage{spconf,amsmath,graphicx}
\usepackage{hyperref}       
\usepackage{url}            
\usepackage{booktabs}       
\usepackage{amsfonts}       
\usepackage{nicefrac}       
\usepackage{xcolor}         
\newcommand{\eg}{\textit{e}.\textit{g}.}
\usepackage{multirow}
\usepackage{pdfpages}
\usepackage{amssymb}
\usepackage{array}
    \makeatletter
    \newcommand{\thickhline}{%
        \noalign {\ifnum 0=`}\fi \hrule height 1pt
        \futurelet \reserved@a \@xhline
}
\usepackage[section]{placeins}

\title{Rethinking Implicit Neural Representations for vision Learners}
\name{Yiran Song$^{1}$, Qianyu Zhou$^{1 \dagger}$ and Lizhuang Ma$^{1 \dagger}$\thanks{$^{\dagger}$Corresponding authors.}}
\address{$^{1}$ Shanghai Jiao Tong University  \\  \{songyiran,zhouqianyu\}@sjtu.edu.cn,   ma-lz@cs.sjtu.edu.cn}

\begin{document}
%
\maketitle

\begin{abstract}
Implicit Neural Representations (INRs) are powerful to parameterize continous signals in computer vision. However, almost all INRs methods are limited to low-level tasks, \emph{e.g.,} image/video compression, super-resolution, and image generation. The questions on how to explore INRs to high-level tasks and deep networks are still under-explored. Existing INRs methods suffer from two problems: 1) narrow theoretical definitions of INRs are inapplicable to high-level tasks; 2) lack of representation capabilities to deep networks. Motivated by above facts, we reformulate the definitions of INRs from a novel perspective, and propose an innovative Implicit Neural Representation Network (INRN), which is the first study of INRs to tackle both low-level and high-level tasks. Specifically, we present three key designs for basic blocks in INRN along with two different stacking ways and corresponding loss functions. Extensive experiments with analysis on both low-level task (image fitting) and high-level vision tasks (image classification, object detection, instance segmentation) demonstrate the effectiveness of the proposed method.

\end{abstract}
\begin{keywords}
Implicit Neural Representations, Computer Vision, Pattern Recognition, Representation Learning
\end{keywords}
\section{Introduction}
\label{sec:intro}


Implicit neural representations (INRs)  \cite{park2019deepsdf, mescheder2019occupancy,chen2019net} have recently shown their superiorities in computer vision, which aims to parameterize the signal as a continuous function that maps
low-dimension spatial/temporal coordinates to the value space. 
Previous INRs approaches mainly focus on low-level tasks, \emph{e.g.,} image super-resolution \cite{chen2021learning, bemana2020x, stanley2007compositional, skorokhodov2021adversarial, anokhin2021image}, image/video compression \cite{stanley2007compositional, chen2021nerv, zhang2021implicit, dupont2021coin}, or image translation \cite{karras2021alias, shaham2021spatially}. Despite its gratifying progress, little attention has been paid to high-level vision tasks, \emph{e.g.,} image classification or segmentation. Existing INRs methods \cite{sitzmann2020implicit, chen2021nerv, mildenhall2020nerf} suffer from two key problems when directly applying to high-level tasks:

\textbf{(i) Existing theoretical definitions of INRs are narrow and not applicable to high-level tasks.}
Previous IRNs approaches \cite{sitzmann2020implicit, tancik2020fourier} only parameterizing the ``seen'' images, however, the images of training and testing sets do not coincide for high-level tasks.
In this paper, we conduct the first pilot study on the question: \textit{how to reformulate the dedinitions of INRs that are suitable for both low-lvel and high-level task?}
Not limited to fitting images into a continuous function, we reformulate INRs as a way to parameterize a high-level object using a neural network. This mapping function $f$ maps a fixed definition domain to a fixed value domain. 

\textbf{(ii) Representation capabilities of existing INRs models are insufficient.}
Most current INRs approaches consist of several MLP layers, \emph{e.g.,} SIREN \cite{sitzmann2020implicit}, LIIF \cite{jiang2020local}, IPF \cite{zhang2021implicit}, and NeRV \cite{chen2021nerv}. Those tiny models lack the representation capability to handle complex tasks and they all perform poorly when the original model is used as a basic block for a deeper network. Thus, the questions on how to extend such INRs methods to deeper networks are important and necessary.

Motivated by above facts, we reformulate the definitions of INRS from a novel perspective, and based on the reformulated definitions of INRs, we present a simple, scalable Implicit Neural Representation (INRe) block with three key designs. Then, stacked on INRe blocks,
we propose a \textbf{Implicit Neural Representation Network} (INRN) , which is the first work of INRs to handle both the low-level and high-level vision tasks. Specifically, we present two stacking ways with the loss functions and training schedules for INRN: single-stage and multi-stage architecture, to deal with both low-level tasks and high-level tasks, respectively. 
To demonstrate the effectiveness of our proposed approach, we provide extensive experiments and analysis on a variety of vision tasks. As shown in Figure \ref{fig:image_fitting}, we illustrate that: $1)$ For low-level tasks, our method achieves superior accuracy compared with previous INRs methods. $2)$ For high-level tasks, our method can produce competitive and promising outcomes. Our contributions  can be summarized into three parts:
\begin{figure*}[htb]
\centering
\includegraphics[width=0.9\linewidth]{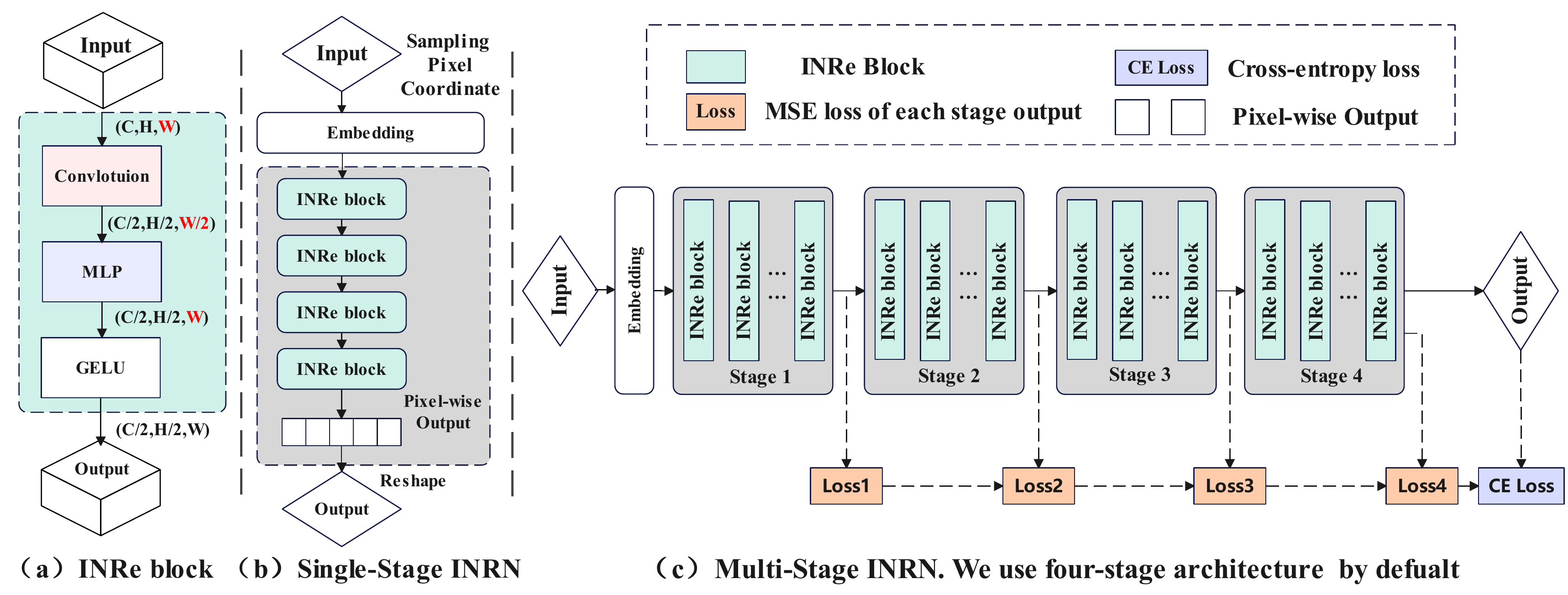}
\vspace{-3mm}
\caption{
(a) INRe block architecture. (b) The single-stage INRN architecture, supervised only by the final outputs. (c) The multi-stage INRN architecture, supervised by multi-stage outputs. The two models are stacked by INRe blocks.
}
\label{fig:INRN}
\end{figure*}

$\bullet$ From a new perspective, we rethink the theoretical definition of IRNs and design a basic block ``INRe'', which has strong expressiveness and generality to handle multiple tasks. 

$\bullet$ We present a new Implicit Neural Representation Network (INRN) stacked by INRe blocks, which is first study to explore INRs for both low-level and high-level vision tasks. We also propose two stacking ways of INRN along with the corresponding loss functions and training pipelines.

$\bullet$ Extensive experiments demonstrate the effectiveness of our method in low-level (image fitting) and high-level ( classification, detection and segmentation) vision tasks.
\section{METHODOLOGY}
\label{sec:format}
\subsection{Rethinking of Implicit Neural Representations}\label{InRe} 


Common INRs were originally defined in the 2D domain as functions $f: \mathbb{R}^{2} \rightarrow \mathbb{R}^{3}$ that parameterize a given image $I$. The pixel coordinates are input $(x, y)$, and the output is the appropriate RGB colors. However, this theoretical definition can only parameterize ``seen'' images, \emph{i. e.}, the same image is used for training and inference, and it cannot directly handle high-level (semantic-level) tasks. The training set and the testing set do not always coincide in high-level tasks, and typical approaches of utilizing pixels to parameterize images do not work well in these cases. 

As a result, we propose a new theoretical definition of INRs : the function $f_{\theta}: \mathbb{R}^{E} \rightarrow \mathbb{R}^{Output}$ is defined to parameterize a high-level object $ \mathcal{D}=\left\{\left(\mathbf{x}_{i}, f\left(\mathbf{x}_{i}\right)\right)\right\}$, where $E$ is the embedding function that maps all sub-elements $\mathbf{x}_{i}$ to a fixed space. The target space of the high-level object determines the output dimension. For low-level tasks, this object $D$ is an image or a video, and its sub-elements are pixels or frames. For high-level tasks, this object $D$ is a dataset, their sub-elements are images, the value domain space is $\mathbb{R}^{N}$, and $N$ is the number of categories. Our goal is to design a network structure that can find the best subset of parameters to approximate the function based on this definition. 

\vspace{-5mm}
\subsection{Implicit Neural Representation (INRe) Block}
In this subsection, we introduce the three core designs of our implicit Neural Representation (INRe) block:

\textbf{Hybrid Architecture.}
We aim to find the mapping function $f$ in the hypothesis space $\mathbb{R}^{N}$ defined by the neural network. 
For small models, $f$ has a significantly higher dimensionality than $\mathbb{R}^{N}$, adding MLP to increase the hypothesis space can considerably increase model expressiveness and attain high performance. However, MLP causes the complexity and parameters rapidly grow when the model size increasing, which suffers greatly from over-fitting. To address this issue, we employ a hybrid architecture to boost the model's generalization. We add a convolutional layer.
The convolution operation adds an inductive error to the model, which helps to avoid over-fitting. Simultaneously, the convolution process shrinks the input in the spatial dimension compared to the original input, reducing the hypothesis space defined by the MLP layer. This method of regulating the model's complexity also aids in reducing over-fitting. 

\textbf{Compression-Expansion Architecture.}
The standard implicit representation approach employs a flat structure, with the number of channels being constant throughout all levels except the first. The channel $C$ is substantially greater than the image's initial dimension (most $3$ channels are expanded to 512 or 768 by default). The flat structure increases the complexity significantly. Furthermore, direct mapping of small parameters to large dimensions in the first layer introduces considerable empirical errors into the model, lowering its accuracy. 
In our INRe block, we adopt a compression-expansion structure rather than the original flat structure. By default, INRe has 2 layers. We compress the input's dimensions in the first layer. This mapping enables the model to extract feature representation from high-dimensional input. We expand the dimensions in the second layer to improve the model's comprehension of the inputs while mapping from low to high dimensions. 

\begin{table}[h!]
\centering
\caption{\textbf{Different INRe architectures}. TS: Training speed. Para.: model parameters. ET: Encoding Time. We adjust the training epoch of each method so that the encoding time (total training time) can be compared.}
\resizebox{\linewidth}{!}{
\begin{tabular}{ccccc}
\hline
    Arthitecture & Para.(M) & \begin{tabular}{c} TS\end{tabular}& \begin{tabular}{c} ET\end{tabular}  & PSNR \\ 
\hline
    Only MLP & 3.54   &  2.9$\times$   & 1.0$\times$ & 25.39   \\
    Front Conv + MLP    & 4.35  & 1$\times$    & 3.0$\times$  & 30.83    \\
    Post Conv  + MLP    & 15.34  & 2.9$\times$   & 1.0$\times$ & 30.93  \\
    Ours    & 13.0  & 1.5$\times$   & 1.5$\times$ & 32.13  \\ 
\hline
\end{tabular}
}
\label{Conv compare}
\end{table}
\textbf{Adjusting the Activation Function.} The activation function in previous IRNs approaches was mainly ReLU. ReLU does not solve the gradient explosion problem and may introduce dead ReLU, \emph{i.e.,} while computing the gradient, too many values are below 0, resulting in the majority of the network's components never being updated. As such, we replace ReLU with GELU \cite{hendrycks2016gaussian} in our network.

\subsection{Network Architecture}
\label{Network}
In this subsection, we propose an Implicit Neural Representation Network (INRN) that can be easily scaled up to deep layers. We present two specific architectures for INRN. 

\textbf{Single-Stage Training and Loss Function.}
As shown in Figure \ref{fig:INRN}(b), we use a simple one-stage INRN for low-level tasks. We utilize a combination of L2 loss and Structural Similarity (SSIM) loss as follows:
\begin{equation}
    L=\alpha \frac{1}{N}\left\|f_{\theta}(X)-Y\right\|_{2}+(1-\alpha)\left(1-\operatorname{SSIM}\left(f_{\theta}(X),Y\right)\right)
    \label{eq:6}
\end{equation}
Here $N$ represents the total sample number, $f_{\theta}(X) \in \mathbb{R}^{H \times W \times 3}$ is the INRN prediction reshaped to a image. $Y \in \mathbb{R}^{H \times W \times 3}$ is the ground-truth. $\alpha$ is used to adjust the each losses. 

\textbf{Multi-Stage Training and Loss Function.}
For semantic-level tasks, we use a multi-stage design for better experimental results, as shown in Figure \ref{fig:INRN}(c). We divide the network into several stages. Each stage $S_{i}$ is defined by the parameter $\boldsymbol{\theta}_{i}$, which is represented as follows:
\begin{equation}
    \Phi_{i}: \mathbb{R}^{I_{i}} \rightarrow \mathbb{R}^{O_{i}}, \quad \mathbf{x}_{i} \mapsto \Phi_{i}\left(\mathbf{x}_{i}\right)=\boldsymbol{\theta}_{i}
    \label{eq:8}
\end{equation}
$I_{i}$ and $O_{i}$ are the input and output dimensions of the $ith$ stage. Inspired by \cite{chen2021distilling}, we introduce a knowledge distillation approach to promote model accuracy. Specifically, we use a lightweight convolutional layer to align the scale between the corresponding stage of the teacher network and the student network (INRN). MSE errors of each layer are added to the overall loss function, which is defined as follows:
\begin{equation}
    \mathcal{L}_{ms}=\sum_{i \in \mathbf{I}}MSE\left(\mathcal{T}_{s}^{i}\left(\mathbf{O}_{s}^{i}\right), \mathcal{T}_{t}^{i}\left(\mathbf{O}_{t}^{i}\right)\right)
    \label{eq:9}
\end{equation}
where $T$ represents the transformation that transforms the output to the same scale. $\mathbf{O}_{s}^{i}$ and $\mathcal{T}_{t}^{i}$ denote output results of the $ith$ of the INRN and teacher network. The overall loss function is formulated as follows:
\begin{equation}
    \mathcal{L}_{final}=\lambda_{1}\mathcal{L}_{CE}+\lambda_{2} \mathcal{L}_{ms}
    \label{eq:10}
\end{equation}
where $\mathcal{L}_{CE}$ is the common cross-entropy loss and we use  $\lambda_{1}$ and $\lambda_{2}$ to balance the weight for each loss component.

\begin{table} [h]
\centering
\caption{\textbf{Flat vs Compression-Expansion.} Comp.Coef: compression coefficient. Flat: Flat Architecture. CE: Compression-Expansion. Para: Parameters. Acc.Bef: Accuracy Before, Acc.Aft: Accuracy After.} 
\resizebox{\linewidth}{!}{
\begin{tabular}{cccccc}
\hline
    Methods & \begin{tabular}{c} Comp.\\Coef.\end{tabular}& \begin{tabular}{c} Flat\\Para. \end{tabular}& \begin{tabular}{c} CE \\Para.\end{tabular}  & \begin{tabular}{c} Acc. \\Bef. \end{tabular} &  \begin{tabular}{c} Acc. \\Aft. \end{tabular} \\
\hline
    INRN-S & 0.5 & 2.5M  & 2.86M &  30.53 & 32.12  \\
    INRN-M & 2 & 14M & 5.7M & 65.01 & 64.43    \\
    INRN-L & 2 & 57M & 21M & 75.57& 73.37  \\
\hline
\end{tabular}
}

\label{Flat compare}
\end{table}

\section{EXPERIMENTS}
\label{sec:pagestyle}



\subsection{Main Properties}
We perform a series of experiments to demonstrate our three core design in Section \ref{InRe} are effective. 

\textbf{Conv-MLP architecture.}
We show the comparison results on ``Big Buck Bunn'' dataset. As shown in Table \ref{Conv compare}: 1) only MLPs. 2) convolutional layers added before or after MLPs. 3) Interleaved MLP and convolution. We use the time $t$ of the frames as the input and the frame ($720 \times 1080$ resolution) as the output. The pure MLP has the fewest parameters, but accuracy is much worse than that with convolutional layers. The method with the conv behind the MLP results in a large increase in the number of parameters while not considerably enhancing the model's accuracy. Our method balances precision and parameter number, allowing us to improve precision while preventing much parameter growth.






\textbf{Compression-Expansion Architecture.} 
We use multiple filter widths and model depths to create INRN models of various sizes, named INRN-S, INRN-M, and INRN-L. We set the compression ratio to 2 by default for INRN-M and INRN-L. We use a compression ratio of 0.5 for INRN-S while solving easy assignments. We do tests on the CIFAR-100 dataset (INRN-S) from the ImageNet dataset (INRN-M and INRN-L), as illustrated in Table \ref{Flat compare}. 

\textbf{GELU VS ReLU.}
We validated INRN on multiple tasks such as video compression, classification, and object detection. Compared with ReLU, the GELU activation function exhibited accuracy gains on all of these tasks. 

\textbf{Network Architecture.} 
We separated the model into single-stage and multi-stage stacks for low-level and high-level tasks, respectively, as indicated in \ref{Network}. We utilize a four-stage network for multi-stage networks. As shown in Table \ref{multi-stage}, a design with more blocks in the middle stage and fewer blocks in the two side stages achieves greater accuracy.

\begin{figure}[htb]
\begin{center}
\includegraphics[width=\linewidth]{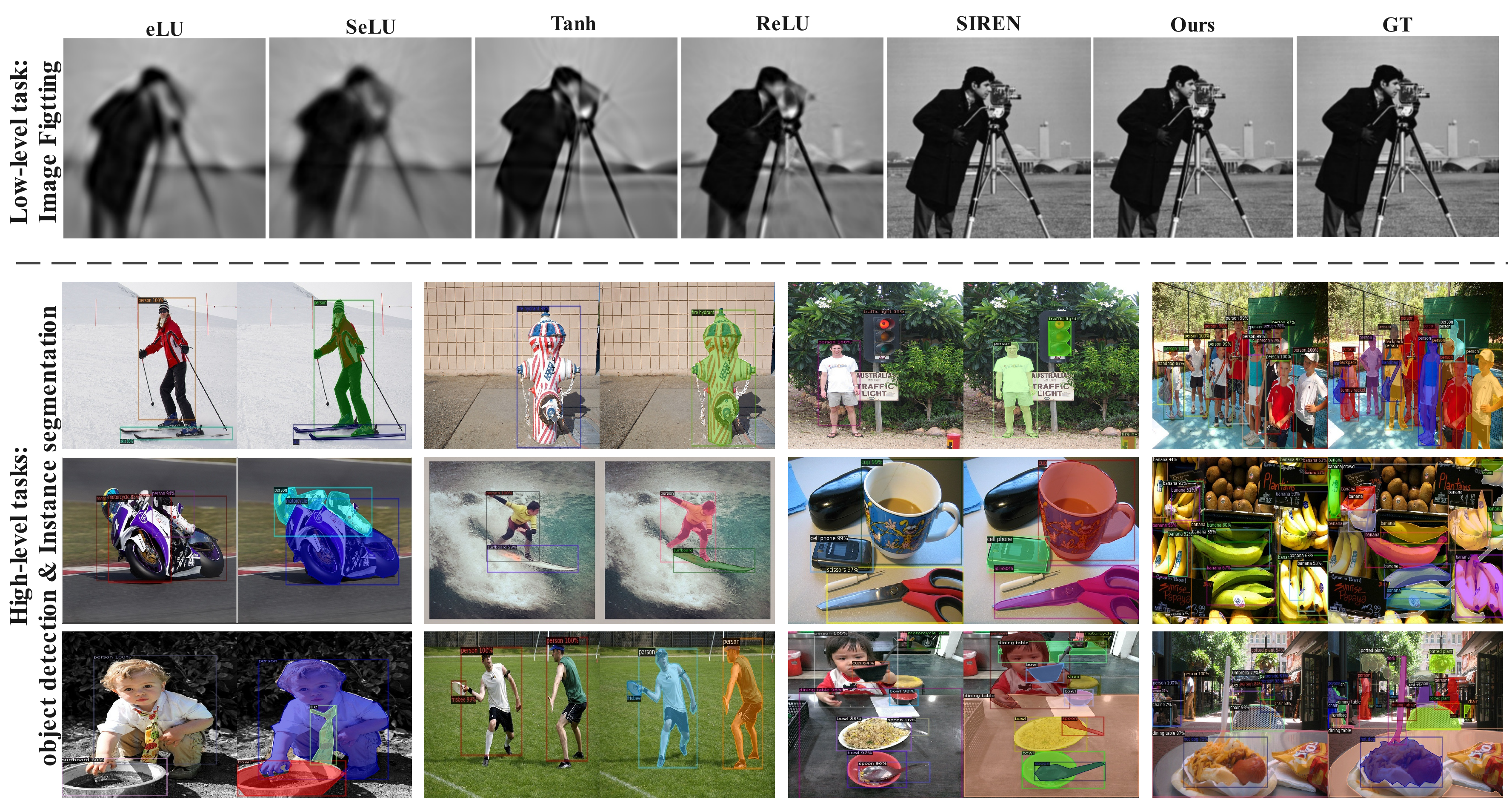}
\end{center}
\vspace{-5mm}
\caption{
\textbf{Visualizations of low-level and high-level tasks.} Top line: comparison of different INRs methods fitting an image. The ground truth is on the left. Bottom line: the results of object detection (odd column) and instance segmentation (even column) on the MS COCO dataset \cite{lin2014microsoft}. 
}
\label{fig:image_fitting}
\end{figure}


\subsection{Main results}
\begin{table} [h!]
\centering
\caption{\textbf{Effect of different stacking methods.} The accuracy based on INRN-M with ImageNet and INRN-L with MS COCO.} 
\resizebox{\linewidth}{!}{
\begin{tabular}{ccccc}
\hline
    Dataset & Model & & Architecture & Accuracy \\
\hline
    \multirow{4}{*}{ImageNet} & \multirow{2}{*}{INRN-M} & Before & [3,3,3,3] & 66.74 \\
     & & After & [2,3,5,2] & 68.43 \\
    \cmidrule {2-5}
     & \multirow{2}{*}{INRN-L} & Before & [3,3,3,3] &  75.71 \\
     & & After & [2,3,4,2] & 76.89 \\
    \hline
    \multirow{2}{*}{COCO} & \multirow{2}{*}{INRN-L} & Before & [3,3,3,3] &  33.42 \\
  && After & [2,3,4,2]  &  38.95 \\
\hline
\end{tabular}
}  
\label{multi-stage}
\end{table}

\textbf{Visualizations of low-level and high-level tasks. }
To get a qualitative sense of the effectiveness of INRN, see Figure \ref{fig:image_fitting}. The visualizations of both low-level and high-level tasks show that our method has achieved outstanding performance.

Compared with prior works, the core innovation of our method is that it can be applied to high-level tasks. We have verified the classification, object detection and segmentation tasks on the CIFAR, Imagenet and COCO datasets.

\textbf{Classification.} There is no implicit neural representation relevant methods in this area and our approach refers to some successes in the knowledge distillation domain. For a fair comparison, we compare our INRN with SOTA methods in this domain, \eg, KD \cite{hinton2015distilling}, AT \cite{zagoruyko2016paying}, CRD \cite{tian2019contrastive}, OFD \cite{heo2019comprehensive}. The KD uses logists information, CRD uses single-layer information, and AT and OFD use multiple-layer information for knowledge distillation. We trained the INRN models on the CIFAR dataset from scratch, with similar scales with other methods, following training pipeline mentioned in Section \ref{Network}. We further conduct experiments on the much bigger ImageNet to verify the scalability of INRN, with ResNet101 as a teacher. All results are summarized in Table \ref{classification}. 

\textbf{Object Detection.}
Similar to the classification task, we use ResNet101 as the teacher network while comparing with existing SOTA methods, \eg, KD \cite{hinton2015distilling} and FitNet \cite{romero2014fitnets}. It should be noted that, other approaches are rely on pre-trained ResNet50. Our algorithm is trained from scratch. This means that our overall training time is much less than other methods, and we obtain competitive accuracy (as shown in Table \ref{Detection}). This proves that our algorithm has extremely strong generalization ability and excellent convergence speed.  

\begin{table} [h!]
\centering
\caption{\textbf{Results on classification.} The accuracy based on INRN-M with ImageNet and INRN-L with COCO datasets.} 
\resizebox{\linewidth}{!}{
\begin{tabular}{ccccccc}
\hline
    Dataset   &     & KD   & AT    & OFD  & CRD   & Ours  \\
\hline
\multirow{2}{*}{CIFAR} & Top-1     & 70.66 & 70.55 & 70.38 & 70.98 & 71.06  \\
& Top-5 & 93.03    &94.51         & 94.09    & 95.01 &  95.19 \\
\hline
\multirow{2}{*}{ImageNet} & Top-1   & 76.66 & 76.69 & 76.81 & 77.17  & 76.70   \\
& Top-5   & 92.68 & 93.81 & 92.78 & 92.93 & 92.94  \\
\hline
\end{tabular}
}  
\label{classification}
\end{table}
\vspace{-7mm}
\begin{table} [h!]
\centering
\caption{\textbf{Results on object detection.} 
}
\resizebox{\linewidth}{!}{
\begin{tabular}{ccccc}
\hline
    \multicolumn{2}{c}{Faster R-CNN} & mAP & AP50 & AP75  \\
    \hline
    Teacher&  +R101-FPN & 42.04 & 62.48 & 45.88   \\
    \hline
    \multirow{4}{*}{Student}& +R50-FPN & 37.93 & 58.84 & 41.05   \\
    & +KD~\cite{hinton2015distilling}  & 38.35 (+0.42) & 59.41 & 41.71  \\
    & +FitNet~\cite{romero2014fitnets} & 38.76 (+0.83) & 59.62 & 41.80 \\
    & +INRN-M & \textbf{39.08 (+1.15)} & \textbf{59.77} & \textbf{42.32}  \\
\hline
\end{tabular}
}
\label{Detection}
\end{table}
\vspace{-7mm}
\begin{table} [h!]
\centering
\caption{\textbf{Results on Instance Segmentation.} }
\resizebox{\linewidth}{!}{
\begin{tabular}{ccccc}
\hline
    \multicolumn{2}{c}{Mask R-CNN} & mAP & AP50 & AP75  \\
    \hline
    Teacher& +R101-FPN & 38.63 & 60.45 & 41.28  \\
    \hline
    \multirow{2}{*}{Student }& +R50-FPN & 35.24 & 56.32 &  37.49  \\
    &  +INRN-M & \textbf{36.52 (+1.28)} & \textbf{57.48} & \textbf{38.01}  \\
\hline
\end{tabular}
} 
\label{Segmentation}
\end{table}
\textbf{Instance Segmentation.}
We reveal the potential of INRN-L for more difficult task, \emph{i.e.,} instance segmentation. To the best of our knowledge, not only is there no relevant implementation in the field of INRs for this task, but also almost no methods in the field of knowledge distillation applied to instance segmentation. As shown in Table \ref{Segmentation}, our algorithm also improves the performance of instance segmentation tasks, notably. We demonstrate that our method can improve the accuracy of existing INRs methods on low-level tasks. Also, on high-level tasks, we can obtain competitive results when compared with classical algorithms in multiple domains. 

\section{CONCLUSION}
\label{sec:typestyle}
In this paper, we rethink implicit neural representations for vision learners and propose an implicit neural representation (INRe) block with three key designs. Besides, we present a simple and scalable Implicit Neural Representation Network stacked by the the INRe blocks, which is the first work to tackle both the low-level and high-level vision tasks via INRs. 
Extensive experiments on multiple vision tasks demonstrate the effectiveness of the proposed method.


\noindent \textbf{Acknowledgements}
This work is supported by National Key Research and Development Program of China (2019YFC1521104), National Natural Science Foundation of China (72192821, 61972157), Shanghai Municipal Science and Technology Major Project  (2021SHZDZX0102), Shanghai Science and Technology Commission (21511101200, 22YF1420300), and Art major project of National Social Science Fund (I8ZD22).
\vfill\pagebreak
\bibliographystyle{IEEEbib}
\bibliography{strings,refs}

\end{document}